\documentclass{article}
\usepackage{spconf}
\usepackage{amsmath}
\usepackage{graphicx}
\usepackage{booktabs}
\usepackage{tabularx}
\usepackage{float}
\usepackage{arydshln}
\usepackage{xspace}
\usepackage{stmaryrd}

\usepackage[backgroundcolor=lightgray,textcolor=red]{todonotes}
\presetkeys{todonotes}{inline}{}

\newcommand{\chain}[0]{$\shortrightarrow$\xspace}

\title{End2End Acoustic to semantic transduction}

\name{
    \begin{tabular}{c}
    Valentin Pelloin$^{1}$ \qquad
    Nathalie Camelin$^{1}$ \qquad
    Antoine Laurent$^{1}$ \qquad
    Renato De Mori$^{2, 3}$ \qquad \\
    Antoine Caubrière$^{1}$ \qquad
    Yannick Estève$^{2}$ \qquad
    Sylvain Meignier$^{1}$
    \end{tabular}
}

\address{
    $^{1}$ LIUM - Le Mans University - France \thanks{This work has been funded by the AISSPER project supported by the French National Research Agency (ANR) under contract ANR-19-CE23-0004-01.} \\
    $^{2}$ LIA - Avignon University - France \\
    $^{3}$ McGill University - Montréal, Canada
}

\begin{document}

\maketitle

\begin{abstract}
In this paper, we propose a novel end-to-end sequence-to-sequence spoken language understanding model using an attention mechanism. It reliably selects contextual acoustic features in order to hypothesize semantic contents. 
An initial architecture capable of extracting all pronounced words and concepts from acoustic spans is designed and tested. With a shallow fusion language model, this system reaches a 13.6 concept error rate (CER) and an 18.5 concept value error rate (CVER) on the French MEDIA corpus, achieving an absolute 2.8 points reduction compared to the state-of-the-art.
Then, an original model is proposed for hypothesizing concepts and their values. This transduction reaches a 15.4 CER and a 21.6 CVER without any new type of context. 
\end{abstract}


\section{Introduction}
    \label{sec:intro}

Spoken Language Understanding (SLU) systems extract semantic contents mentioned in spoken sentences for task-oriented dialogues, question answering, and other conversational applications. 

Semantic contents are fragments of a domain application
ontology that can be defined with a frame language as described in \cite{TurDeMori2011}. 
Frame structures are rich representations of knowledge structures with important properties, such as type models of slot fillers and inheritance of slot properties through chains of frame-slot relations.
With a frame language, a concept is represented as a relation between a \textit{frame name} (e.g. ADDRESS), a frame property, often called \textit{slot}, (e.g. city), and a frame \textit{value}, often called \textit{slot filler} (e.g. Paris).
Properties can be inherited through a \textit{chain of slots}.
Using this semantic representation, speech acts such as \textit{request} can be defined as frames with slots filled by instances of frame structure fragments. For the sake of simplicity, in the following, slots and slot fillers are denoted as \emph{concept} and \emph{value}.%

Concepts can be annotated with word sequences called \emph{supports}. 
Supports can be instantiated with hypotheses obtained by an Automatic Speech Recognition (ASR) system as described in \cite{hakkani2016multi, Zhang2016, Liu2016, simonnet2017} or by direct transduction that generates instances of frame structures from acoustic features.

Supports are word spans expressed by a span of acoustic features. Such a span may have variable duration and depends on the speaker and other factors.
If a value is mentioned, then it is often included in a support. Spans of acoustic feature supports may not be sufficient for unambiguously mentioning a concept instance, even if they contain relevant semantic features. This equivocation can be reduced by selecting relevant spans of acoustic feature contexts using an attention mechanism.

Human experience in understanding spoken sentences, especially in foreign languages, provides evidence that concepts can be hypothesized from spoken language tokens without thinking about, or even knowing, the exact spelling of some relevant words. This motivates the use of a model based on a transduction of acoustic into semantic content supports. 
In this paper, an end-to-end (E2E) SLU architecture is proposed in order to hypothesize fragments of frame instances with acoustic features. It is applied on user turns of the French MEDIA corpus, corresponding to a complex negotiation dialogue task (see Section \ref{sec:datasets}).
This task is particularly difficult since a dialogue turn may contain a large variety of concept-value instances, including repetitions and self-corrections with specific semantic relations between these contents. 

A novelty of the proposed model is the introduction of an attention mechanism that selects contextual acoustic features in order to hypothesize directly concept symbols and their values. 
An initial architecture is first designed and tested in order to hypothesize words and concepts. %
An enriched architecture is then proposed in order to hypothesize concepts and delimiting their supports of underlying acoustic representations while limiting the generation of character hypotheses only for values. As a consequence, no human prior-knowledge is needed to extract values from words of the concept support. %

The paper is organized as follows. Section \ref{sec:related_work} presents related works in the SLU domain. Section \ref{sec:format} presents our E2E SLU architecture, while corpora, experiments and results are described in Section \ref{sec:exps}. Finally, conclusions and perspectives are given in Section \ref{sec:conclusions}.

\section{Related work}
    \label{sec:related_work}

Early speech understanding systems based on natural language semantic parsers and artificial intelligence approaches to beam search and knowledge representation are reviewed in \cite{Klatt1977}. Further developments stressing comparisons of different methods using a common annotated corpus, new parsing techniques and statistical models are reviewed in \cite{Kuhn1995}.
Later, advanced spoken language understanding systems with advanced statistical models and paradigms in various types of applications are described in \cite{TurDeMori2011}. 
More recently, deep neural networks (DNN) for generating semantic domain, intent and slot filler hypotheses were proposed using a pipeline of automatic speech recognition and SLU models. These architectures provided better or comparable results to those obtained with  previous architectures. Examples can be found in \cite{hakkani2016multi, Zhang2016, Liu2016}. Result examples and comparison using SLU in negotiation dialogues are reported in \cite{simonnet2017}. 
Examples of attempts to integrate ASR and SLU functions in  end-to-end (E2E) compact trainable DNN architectures are proposed in \cite{Qian2018, Serdyuk2018, Price2020, Haghani2019, Tomashenko2020, Wang2020}.

In this paper, a novel E2E SLU system is introduced and tested on the French MEDIA corpus. It is motivated by the evidence that the perception of spoken semantic entities and relations is associated with the perception of word spans expressing semantic properties and their values. A basic architecture is introduced with attention on suitable acoustic contexts having relevance dependent on decoder semantic representations. Then, a novel architecture is proposed in order to detect acoustic spans expressing lexical features for clue words of semantic entities and their values. This introduction is motivated by explaining, with prior knowledge, frequent errors observed in the annotated development set. 

Recently, methods have been proposed to learn high-level representations from surface acoustic features for speech-to-speech translation \cite{Baevski2020, Chorowski2019, Chung2020, Khurana2020}.

In this paper, we aim to use high-level representations from surface acoustic features for speech-to-semantic transduction.

In \cite{He2020}, relations in semantic knowledge graphs have been used for inferring answers in question answering dialogues.

In this paper, application domain model relations, explicit or inferred by inheritance, are considered for selecting useful context for the interpretation of acoustic spans of domain relevant concepts.

\section{E2E SLU architecture}
    \label{sec:format}

In this work, we use an encoder-decoder neural network. An attention mechanism as described by \cite{bahdanau2016neural} was designed in order to align the input with the output. Hypothesizing spoken phonemes with an ASR system may be difficult if a training set of phonemes aligned with speech is not available. For this reason, a first attempt using experience on character recognition is used in this paper. Acoustic to concept hypothesizing through to phoneme is left for future work.

The input of the network are 40 dimensional MelFBanks extracted with a Hamming window of 25 ms and 10 ms strides. %
The system output is a sequence of characters predicting words, normalized values, or concept labels. Note that each concept label is represented by one special character (\textit{e.g.} \texttt{ǵ} for the concept \textit{hotel-services}).

\begin{figure}[ht!]
        \centering
        \includegraphics{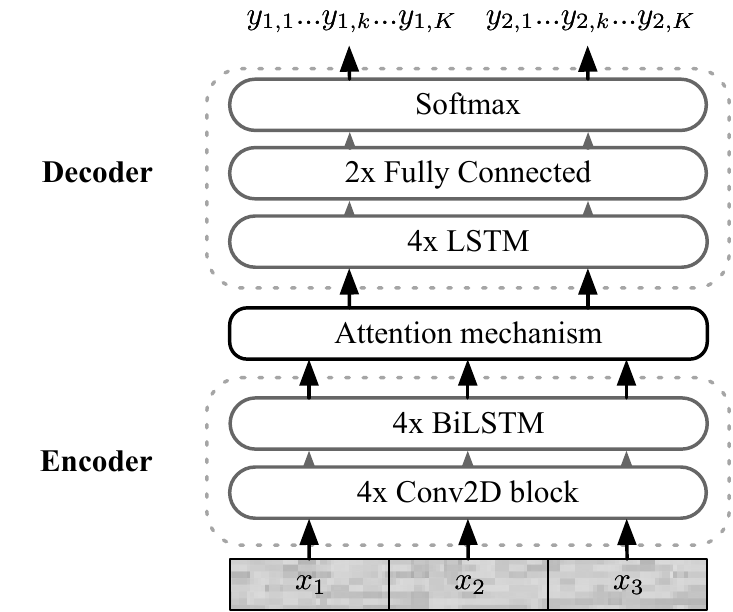}
        \caption{E2E Encoder-Decoder SLU Model}
        \label{fig:e2e_model}
    \end{figure}

Our attention based encoder-decoder architecture, depicted in Figure \ref{fig:e2e_model}, is based on the \emph{Espresso} recipe initially developed for the WSJ ASR task, as described in \cite{Wang2019}.

Let $X = (x_1, ..., x_{T_x})$ be the vector of the input features and $Y = (y_1, ..., y_{T_y})$ be the outputs of the model. Our model computes the outputs $Y$ from an input sequence $X$.
The encoder first uses 4 2-dimensional convolutional blocks (each convolution layer is followed by a batch norm). Then, 4 biLSTM layers are used to obtain the encoder hidden states as:
\begin{equation}
    \text{Encoder}(X) = (\overleftrightarrow{h_1}, ..., \overleftrightarrow{h_{T_x}})
\end{equation}
The decoder uses 4 LSTM followed by 2 fully-connected layers and a softmax. 
The decoder states are computed using the attention mechanism by aligning the encoder hidden states and previous decoder hidden states, as:
\begin{equation}
    Y = \text{Decoder}(H^\text{att})
\end{equation}
The attention mechanism is applied as described in \cite{bahdanau2016neural}:
\begin{align}
    H^\text{att} &= (H^\text{att}_1, ..., H^\text{att}_{T_y}) \\
    H^\text{att}_i &= \sum_j^{T_x} \alpha_{ij}~\overleftrightarrow{h_j} \quad \text{with} \quad \alpha_{ij} = \frac{exp(e_{ij})}{\sum_{k=1}^{T_x} exp(e_{ik})}
\end{align}

$e_{ij}$ is the relevance of the acoustic representation $\overleftrightarrow{h_j}$ and the $i^\text{th}$ semantic output. %
Note that the encoder vectors $\overleftrightarrow{h}$ are computed using only the information from the top-encoder hidden states.

In order to improve the encoder-decoder performance, we use an additional LSTM-based Language Model (LM)  %
that uses \textit{look-ahead} word probabilities to obtain both word and character probabilities (\cite{Wang2019, Hori2019}).

\textit{Shallow-fusion} \cite{gulcehre2015using} score combination is performed in order to take into account the output of the encoder-decoder SLU system and the LSTM-based LM. During the decoding phase with the beam search, hypotheses are scored using the weighted sum of both the encoder-decoder model and the LM scores.

\section{Experiments}
    \label{sec:exps}

    \subsection{Datasets}
        \label{sec:datasets}
        Our architecture is pretrained for ASR and then adapted to SLU. Training is detailed in the next section while we briefly describe the different corpora in this section.
        
        For the transcription model, we use several French generic broadcast news corpora described in \cite{Caubriere2019}, in addition to the training sets of in-domain SLU telephone conversation corpora: PORTMEDIA \cite{Portmedia2012}, MEDIA \cite{Media2004} and DECODA \cite{Decoda2012}. As a result, we obtained a training corpus of 414 hours for ASR. 
        Such a model needs a lot of audio data along with their manual transcription even if a frame acoustic alignment with words is not needed. The system will train itself to align each output character with input acoustic features thanks to its attention mechanism \cite{Chorowski2014, Chan2016, Bahdanau2016}. 

        For the SLU models, only the 24 hours of MEDIA (M) and PORTMEDIA (PM) training sets are used during training.
        These two corpora contain both telephone conversations between two humans in the \emph{Wizard of Oz} mode: one plays a computer, the other a user who wants to obtain information or make a reservation. MEDIA contains requests and lodging booking in France and PORTMEDIA contains conversations about theater shows during the french Avignon Festival. All conversations are manually transcribed and user turns are semantically annotated with concept labels, their supports and the corresponding values. Considering the Table \ref{tab:models_and_outputs} example, the concept label \textit{link-coref} is supported by the word \textit{that} annotated with the normalized value \textit{singular}.
        
        Turns of these datasets contain word spans expressing semantic content mentions mixed with other spans that do not express application domain contents. For this task, user turns may express concepts such as logical AND OR connectors, references to previously mentioned concepts and mentions to entities relative to other entities. Furthermore, system turns tend to solicit the user to provide useful information for composing a \textit{state of the world} to obtain a frame structure in an application domain database.
        
        In the experiments section, we will present SLU results on the MEDIA corpus,  split accordingly to the official training, development (\emph{Dev}) and test sets, containing respectively 727, 79 and 208 dialogues and 13k, 1.2k and 3.6k user utterances. User transcripts are annotated using 76 semantic concepts, along with their supports and the corresponding normalized value. 
\vspace{-0.1cm}       
    \subsection{Training and outputs}
        \label{sec:training_outputs}

        Inspired by the \textit{curriculum-based transfer learning} procedure for SLU proposed in \cite{Caubriere2019}\footnote{Unlike \cite{Caubriere2019}, we did not choose to train our model on the Named Entity Recognition (NER) task.}, we first train a model for the ASR downstream task. For that purpose, we use out-of-domain and in-domain ASR data (as described in the previous section), and then fine-tune using our in-domain corpora. This first model is denoted as \emph{ASR} model.
        
        Thereafter, the model is retrained for the SLU task using all the in-domain SLU data (MEDIA and PORTMEDIA train sets, denoted M+PM), and then fine-tuned with the target MEDIA data.
        The transition from ASR to SLU is made by enriching the output with concept symbols. 
       In contrast with \cite{Caubriere2019, Ghannay2018}, the last fully connected layer is just extended.  As a consequence, the model benefits entirely from its previous training. 
       This second model is denoted as \emph{AllWords-C} model.

        \cite{Ghannay2018} showed that hypothesizing only words belonging to concept support enhanced the SLU performance evaluation. Thus, we introduce a third model that adapts the \emph{AllWords-C} model on the specific outputs: concepts and words from supports. A new character is then added to the last layer in order to output a global character * instead of all out-of-concept-support words. 
        This third \emph{SupWords-C} model is obtained with transfer learning on these outputs.
        
       Finally, a novel solution is introduced to perform a direct transduction from acoustic features to concept-value pairs. 
       To do so, we re-adapt a last time the \emph{AllWords-C} model in order to output directly the normalized values and the concept. In that case, characters no longer represent pronounced words. They represent either one concept or one character of a normalized value.
        This model is denoted as \emph{NormValues-C} model. 
        In that last architecture, frame structure fragments are obtained with a fully automatic E2E sequence 2 sequence architecture.
       
        All models and their output formats are summarized in Table \ref{tab:models_and_outputs}. Note that, for the sake of clarity, concepts are written in full instead of their special character output.

\newcolumntype{L}{>{\centering\arraybackslash}p{0.95\linewidth}}        

\begin{table}[ht!]
\centering
\begin{tabular}{L} 
\hline
\textit{\textbf{ASR}} \\
\multicolumn{1}{c}{ ASR \chain ASR M+PM} \\
\footnotesize{\texttt{is there a swimming-pool in that one}} \\ 
\hline
\textbf{\textit{AllWords-C}} \\
\multicolumn{1}{c}{ ASR model \chain SLU M+PM \chain SLU M} \\
\footnotesize{\texttt{is there a $<$hotel-services$>$ swimming-pool $<$/hotel-services$>$ in $<$linkref-coref$>$ that $<$/linkref-coref$>$ $<$objectbd$>$ one $<$/objectbd$>$}} \\ 
\hline
\textbf{\textit{SupWords-C}} \\
\multicolumn{1}{c}{\textit{AllWords-C} model \chain SLU M*} \\
\footnotesize{\texttt{* $<$hotel-services$>$ swimming-pool $<$/hotel-services$>$ * $<$linkref-coref$>$ that $<$/linkref-coref$>$ $<$objectbd$>$ one $<$/objectbd$>$}} \\ 
\hline
\textbf{\textit{NormValues-C}} \\
\multicolumn{1}{c}{\textit{AllWords-C} model \chain SLU Norm M*} \\
\footnotesize{\texttt{* $<$hotel-services$>$ swimming-pool $<$/hotel-services$>$ * $<$linkref-coref$>$ singular $<$/linkref-coref$>$ $<$objectbd$>$ hotel $<$/objectbd$>$}} \\
\hline
\end{tabular}
    \caption{Model (chain of training) and output for each proposed configuration, based on the user utterance "Is there a swimming-pool in that one?".}
    \label{tab:models_and_outputs}
\end{table}

    \subsection{Evaluation Protocol}
The architecture shown in Figure \ref{fig:e2e_model} is used in order to generate outputs with different components, all evaluated according to the concept error rate (CER) and concept-value error rate (CVER) metrics. Insertion, deletion, and substitution errors are used to compute evaluation measures and to perform the error analysis.

Normalized values, \textit{i.e.} the slot fillers, needed for the evaluation in CVER,  are obtained with a set of manually designed regular expressions in \emph{AllWords-C} and \emph{SupWords-C} configurations. Expressions are applied to the outputs of each concept support, as in \cite{Tomashenko2020,Caubriere2019,Ghannay2018,simonnet2017}. 
Using the \emph{NormValues-C} outputs, these handmade rules are not necessary anymore.

    \subsection{Results}

Table \ref{tab:results-media} summarizes the results obtained with the three considered architecture solutions, without and with LM, on the MEDIA corpus on both the Dev and Test sets.

\setlength\dashlinedash{0.2pt}
\setlength\dashlinegap{1.5pt}

\begin{table}[!ht]
\centering
\begin{tabular}{@{}rcccc@{}}
\toprule
& \multicolumn{2}{c}{\textbf{Dev}} & \multicolumn{2}{c}{\textbf{Test}} \\
\textbf{\%} & \textbf{CER} & \textbf{CVER} & \textbf{CER} & \textbf{CVER} \\ \midrule
\multicolumn{5}{c}{\textbf{Without a Language Model}} \\\hline
\emph{AllWords-C} \cite{Caubriere2019} & -- & -- & 21.6 & 27.7 \\\hdashline
\emph{AllWords-C}                      & 18.1 & 22.5 & 15.6 & \textbf{20.4} \\
\emph{SupWords-C}                      & 17.3 & 22.0 & 15.6 & 20.5 \\
\emph{NormValues-C}              & 16.0 & 21.9 & \textbf{15.4} & 21.7 \\\midrule
\multicolumn{5}{c}{\textbf{With a Language Model}} \\\hline
\emph{AllWords-C} \cite{Caubriere2019} & -- & -- & 18.1 & 22.1 \\
\emph{SupWords-C} \cite{Caubriere2019} & -- & -- & 16.4 & 20.9 \\\hdashline
\emph{AllWords-C}                      & 16.1 & 20.4 & \textbf{13.6} & \textbf{18.5} \\
\emph{SupWords-C}                      & 17.6 & 22.5 & 15.5 & 20.5 \\
\emph{NormValues-C}              & 16.1 & 22.0 & 15.4 & 21.6 \\
\bottomrule
\end{tabular}
\caption{Results obtained on the MEDIA Dev and Test Corpora by our models compared to the State of the art.} 
\label{tab:results-media}
\end{table}
Without LM, the best CER results are obtained with \emph{NormValues-C} model, both for Dev and Test. This supports the conjecture that concepts can be perceived without considering orthographic word  transcriptions. As we could expect, when CER are similar, best CVER are achieved using human rules. Nevertheless, very encouraging results are observed, notably on Dev, when the CER is lower than others.
For the sake of comparison, \cite{Caubriere2019} state-of-the-art results related to the use of an E2E approach without LM, are reported with 21.6\% CER and 27.7\% CVER, showing an absolute gain of 6 points.

Unlike in \cite{Caubriere2019} where best results are obtained with the \emph{SupWords-C} configuration, our architecture does not benefit from this kind of representation. We suppose that this is due to the architectural difference: Attention Mechanism (AM) \textit{vs.} biLSTM/CTC. 
In the latter, the star is used as a correspondence to the input acoustic frame when predicting outside-of-concept characters. %
In our AM system, such a star is not necessarily needed, as the output sequence is not constrained to have the same length as the input. The AM is designed to directly select relevant acoustic spans.

Using word hypotheses and LM, the best 13.6\% CER and 18.5\% CVER are observed for \emph{AllWords-C}, showing a gain of 2 points using the LM and 2.8 points CER considering the best \cite{Caubriere2019} state-of-the-art result even without significant contribution of the LM on  %
\emph{NormValues-C} and \emph{SupWords-C}. 
This can be explained by the LM weight, optimized on the Dev set, which is close to zero. This needs further investigation.

A first analysis of detailed concept errors on Dev shows that the most frequent ones are insertions and deletions of logical connectors and co-references. %

\section{Conclusions and perspectives}
    \label{sec:conclusions}

An E2E SLU architecture is introduced, based on an encoder-decoder model
with attention mechanism focusing on acoustic representations, useful for generating hypotheses of words, concepts or values. Several combinations have been investigated using or not LM reaching better results than state-of-the-art. %

The results indicate that word knowledge can be a relevant context for semantic interpretation, particularly if it is selected by an appropriate attention mechanism. %
The solutions evaluated in this paper are based either on word or value representations
using characters. 
In future work we also plan to use two decoders, one that outputs each character (\emph{allWords-C}) and the other one which focuses on concept-value pairs. This could, at least, improve our CVER without degrading the CER using the \emph{NormValues-C} representation.
Finally, new types of LM including semantic hypotheses will also  be investigated in future work.

\vfill\pagebreak

    \label{sec:refs}

\small
    \bibliographystyle{IEEEbib}
    \bibliography{bibliography}

\begin{thebibliography}{10}

\bibitem{TurDeMori2011}
G.~Tur and R.~De~Mori,
\newblock ``Chapter 1: Spoken language understanding for human/machine
  interactions,''
\newblock in {\em Spoken Language Understanding: Systems for Extracting
  Semantic Information from Speech}. 2011.

\bibitem{hakkani2016multi}
D.~Hakkani-T{\"u}r, G.~T{\"u}r, A.~Celikyilmaz, Y.~N. Chen, J.~Gao, L.~Deng,
  and Y.~Y. Wang,
\newblock ``Multi-domain joint semantic frame parsing using bi-directional
  rnn-lstm.,''
\newblock in {\em INTERSPEECH}, 2016, pp. 715--719.

\bibitem{Zhang2016}
X.~Zhang and H.~Wang,
\newblock ``A joint model of intent determination and slot filling for spoken
  language understanding,''
\newblock in {\em IJCAI}, 2016, p. 2993–2999.

\bibitem{Liu2016}
B.~Liu and I.~Lane,
\newblock ``Attention-based recurrent neural network models for joint intent
  detection and slot filling,''
\newblock in {\em INTERSPEECH}, 2016, pp. 685--689.

\bibitem{simonnet2017}
E.~Simonnet, S.~Ghannay, N.~Camelin, Y.~Est{\`e}ve, and R.~De~Mori,
\newblock ``{ASR error management for improving spoken language
  understanding},''
\newblock in {\em {INTERSPEECH}}, 2017.

\bibitem{Klatt1977}
D.~H. Klatt,
\newblock ``Review of the {ARPA} speech understanding project,''
\newblock {\em The Journal of the Acoustical Society of America}, vol. 62, pp.
  1345, 1977.

\bibitem{Kuhn1995}
R.~{Kuhn} and R.~{De Mori},
\newblock ``The application of semantic classification trees to natural
  language understanding,''
\newblock {\em TPAMI 17}, pp. 449--460, 1995.

\bibitem{Qian2018}
Y.~Qian, R.~Ubale, V.~Ramanaryanan, P.~Lange, D.~Suendermann-Oeft, K.~Evanini,
  and E.~Tsuprun,
\newblock ``{Exploring ASR-free end-to-end modeling to improve spoken language
  understanding in a cloud-based dialog system},''
\newblock {\em ASRU}, pp. 569--576, 2017.

\bibitem{Serdyuk2018}
D.~Serdyuk, Y.~Wang, C.~Fuegen, A.~Kumar, B.~Liu, and Y.~Bengio,
\newblock ``{Towards End-to-end Spoken Language Understanding},''
\newblock {\em ICASSP}, pp. 5754--5758, 2018.

\bibitem{Price2020}
R.~Price, M.~Mehrabani, and S.~Bangalore,
\newblock ``{Improved End-To-End Spoken Utterance Classification with a
  Self-Attention Acoustic Classifier},''
\newblock {\em ICASSP}, pp. 8504--8508, 2020.

\bibitem{Haghani2019}
P.~Haghani, A.~Narayanan, M.~Bacchiani, G.~Chuang, N.~Gaur, P.~Moreno,
  R.~Prabhavalkar, Z.~Qu, and A.~Waters,
\newblock ``{From Audio to Semantics: Approaches to End-to-End Spoken Language
  Understanding},''
\newblock {\em SLT}, pp. 720--726, 2018.

\bibitem{Tomashenko2020}
N.~Tomashenko, C.~Raymond, A.~Caubriere, R.~De~Mori, and Y.~Est{\`e}ve,
\newblock ``{Dialogue History Integration into End-to-End Signal-to-Concept
  Spoken Language Understanding Systems},''
\newblock {\em ICASSP}, pp. 8509--8513, 2020.

\bibitem{Wang2020}
P.~Wang, L.~Wei, Y.~Cao, J.~Xie, and Z.~Nie,
\newblock ``Large-scale unsupervised pre-training for end-to-end spoken
  language understanding,''
\newblock {\em ICASSP}, pp. 7994--7998, 2020.

\bibitem{Baevski2020}
A.~Baevski, S.~Schneider, and M.~Auli,
\newblock ``vq-wav2vec: Self-supervised learning of discrete speech
  representations,''
\newblock in {\em ICLR}, 2020.

\bibitem{Chorowski2019}
J.~{Chorowski}, R.~J. {Weiss}, S.~{Bengio}, and A.~{van den Oord},
\newblock ``Unsupervised speech representation learning using wavenet
  autoencoders,''
\newblock {\em TASLP}, pp. 2041--2053, 2019.

\bibitem{Chung2020}
Y.~A. Chung and J.~Glass,
\newblock ``Improved speech representations with multi-target autoregressive
  predictive coding,''
\newblock in {\em ACL}, 2020, pp. 2353--2358.

\bibitem{Khurana2020}
S.~Khurana, A.~Laurent, and J.~Glass,
\newblock ``Cstnet: Contrastive speech translation network for self-supervised
  speech representation learning,'' 2020,
\newblock arXiv:2006.02814.

\bibitem{He2020}
Z.~{He}, Y.~{He}, Q.~{Wu}, and J.~{Chen},
\newblock ``Fg2seq: Effectively encoding knowledge for end-to-end task-oriented
  dialog,''
\newblock in {\em ICASSP}, 2020, pp. 8029--8033.

\bibitem{bahdanau2016neural}
D.~Bahdanau, K.~Cho, and Y.~Bengio,
\newblock ``Neural machine translation by jointly learning to align and
  translate,''
\newblock in {\em ICLR}, 2015.

\bibitem{Wang2019}
Y.~{Wang}, T.~{Chen}, H.~{Xu}, S.~{Ding}, H.~{Lv}, Y.~{Shao}, N.~{Peng},
  L.~{Xie}, S.~{Watanabe}, and S.~{Khudanpur},
\newblock ``Espresso: A fast end-to-end neural speech recognition toolkit,''
\newblock in {\em ASRU}, 2019, pp. 136--143.

\bibitem{Hori2019}
T.~Hori, J.~Cho, and S.~Watanabe,
\newblock ``{End-to-end Speech Recognition with Word-Based Rnn Language
  Models},''
\newblock in {\em SLT}, 2018, pp. 389--396.

\bibitem{gulcehre2015using}
C.~Gulcehre, O.~Firat, K.~Xu, K.~Cho, L.~Barrault, H.~C. Lin, F.~Bougares,
  H.~Schwenk, and Y.~Bengio,
\newblock ``On using monolingual corpora in neural machine translation,'' 2015,
\newblock arXiv:1503.03535.

\bibitem{Caubriere2019}
A.~Caubri{\`{e}}re, N.~Tomashenko, A.~Laurent, E.~Morin, N.~Camelin, and
  Y.~Est{\`{e}}ve,
\newblock ``{Curriculum-based transfer learning for an effective end-to-end
  spoken language understanding and domain portability},''
\newblock {\em INTERSPEECH}, pp. 1198--1202, 2019.

\bibitem{Portmedia2012}
F.~Lef{\`e}vre, D.~Mostefa, L.~Besacier, Y.~Est{\`e}ve, M.~Quignard,
  N.~Camelin, B.~Favre, B.~Jabaian, and Lina~M. Rojas~B.,
\newblock ``{Leveraging study of robustness and portability of spoken language
  understanding systems across languages and domains: the PORTMEDIA corpora},''
\newblock in {\em LREC}, 2012.

\bibitem{Media2004}
L.~Devillers, H.~Maynard, S.~Rosset, P.~Paroubek, K.~McTait, D.~Mostefa,
  K.~Choukri, L.~Charnay, C.~Bousquet, N.~Vigouroux, F.~B{\'e}chet, L.~Romary,
  J.~Y. Antoine, J.~Villaneau, M.~Vergnes, and J.~Goulian,
\newblock ``The {F}rench {MEDIA}/{EVALDA} project: the evaluation of the
  understanding capability of spoken language dialogue systems,''
\newblock in {\em LREC}, 2004.

\bibitem{Decoda2012}
F.~Bechet, B.~Maza, N.~Bigouroux, T.~Bazillon, M.~El-Beze, R.~De Mori, and
  E.~Arbillot,
\newblock ``Decoda: a call-centre human-human spoken conversation corpus,''
\newblock in {\em LREC}, 2012, pp. 1343--1347.

\bibitem{Chorowski2014}
J.~Chorowski, D.~Bahdanau, K.~Cho, and Y.~Bengio,
\newblock ``End-to-end continuous speech recognition using attention-based
  recurrent nn: First results,''
\newblock in {\em NIPS}, 2014.

\bibitem{Chan2016}
W.~Chan, N.~Jaitly, Q.~V. Le, and O.~Vinyals,
\newblock ``Listen, attend and spell: A neural network for large vocabulary
  conversational speech recognition,''
\newblock in {\em ICASSP}, 2016.

\bibitem{Bahdanau2016}
D.~{Bahdanau}, J.~{Chorowski}, D.~{Serdyuk}, P.~{Brakel}, and Y.~{Bengio},
\newblock ``End-to-end attention-based large vocabulary speech recognition,''
\newblock in {\em ICASSP}, 2016, pp. 4945--4949.

\bibitem{Ghannay2018}
S.~Ghannay, A.~Caubri{\`e}re, Y.~Est{\`e}ve, N.~Camelin, E.~Simonnet,
  A.~Laurent, and E.~Morin,
\newblock ``{End-to-end named entity and semantic concept extraction from
  speech},''
\newblock in {\em SLT}, Athens, Greece, 2018.

\end{thebibliography}

\end{document}